\documentclass{article} %
\usepackage{iclr2025_conference,times}

\usepackage{amsmath,amsfonts,bm}

\def\eqref#1{equation~\ref{#1}}

\def\1{\bm{1}}

\DeclareMathAlphabet{\mathsfit}{\encodingdefault}{\sfdefault}{m}{sl}
\SetMathAlphabet{\mathsfit}{bold}{\encodingdefault}{\sfdefault}{bx}{n}

\usepackage{hyperref}
\usepackage{url}
\usepackage{amssymb}
\usepackage[ruled,vlined,linesnumbered]{algorithm2e}
\usepackage{booktabs}
\usepackage{graphicx}
\usepackage{wrapfig}

\title{Bourbaki: Self-Generated and Goal-Conditioned MDPs for Theorem Proving}

\author{%
  Matthieu Zimmer\thanks{Equal contribution} \\
  Huawei Noah's Ark Lab \\
  \And
  Xiaotong Ji \footnotemark[1] \\
  Huawei Noah's Ark Lab\\
  Imperial College London\\
  \And
  Rasul Tutunov \thanks{Equal second author}\\
  Huawei Noah's Ark Lab\\
  \And
  Anthony Bordg \footnotemark[2] \\
  Huawei Lagrange Center \\
  \And
  Jun Wang \\
  UCL Centre for AI \\
  \And
  Haitham Bou Ammar \thanks{Correspondence to: haitham.ammar@huawei.com}\\
  Huawei Noah's Ark Lab \\
  UCL Centre for AI \\
}

\iclrfinalcopy %
\begin{document}

\maketitle

\begin{abstract}
Reasoning remains a challenging task for large language models (LLMs), especially within the logically constrained environment of automated theorem proving (ATP), due to sparse rewards and the vast scale of proofs. These challenges are amplified in benchmarks like \texttt{PutnamBench}, which contains university-level problems requiring complex, multi-step reasoning. To address this, we introduce self-generated goal-conditioned MDPs (sG-MDPs), a new framework in which agents generate and pursue their subgoals based on the evolving proof state. Given this more structured generation of goals, the resulting problem becomes more amenable to search. We then apply Monte Carlo Tree Search (MCTS)-like algorithms to solve the sG-MDP, instantiating our approach in Bourbaki (7B), a modular system that can ensemble multiple 7B LLMs for subgoal generation and tactic synthesis. On \texttt{PutnamBench}, Bourbaki (7B) solves 26 problems, achieving new state-of-the-art results with models at this scale.
\end{abstract}

\section{Introduction}
Reasoning is one of the most important skills in human intelligence, 
and building machines that can reason is a key goal in artificial intelligence. 
A central task in this area is automated theorem proving: automatically generating formal proofs for mathematical statements. Theorem proving is not only useful in mathematics, where it helps create precise and verifiable proofs, but also in fields like formal verification, where it can prove the correctness of complex systems.
Because it requires precise, multi-step logical thinking, theorem proving is a strong benchmark for testing how well AI systems can reason. While large language models (LLMs) have shown surprising abilities in math and logic, they often struggle to produce complete and reliable proofs in a structured way.

A wide range of approaches have been proposed for automated theorem proving (ATP) in formal systems such as Lean, Isabelle, and Coq. Symbolic methods, including saturation-based provers and tactic-based automation, offer strong guarantees but struggle with the combinatorial complexity of creative problems. To address this, neural models have been introduced to guide search, predict tactics, or select useful premises from large libraries, leading to higher success rates for automated theorem proving in formal libraries. More recently, large language models (LLMs) have shown promise in generating proof steps by drawing on broad mathematical knowledge. However, LLMs alone often hallucinate or fail to produce mathematical proofs. Therefore, their integration with formal systems for machine-checkable proofs, like those underpinning proof assistants, is essential. This has led to hybrid approaches that combine LLMs with search algorithms, such as best-first or Monte Carlo Tree Search (MCTS) or with reinforcement learning, resulting in state-of-the-art performance on benchmarks like MiniF2F and FIMO. 

Yet progress stalls on harder problems, particularly those in the new PutnamBench dataset~ \citep{tsoukalas2024putnambench}, which formalises hundreds of problems from the challenging William Lowell Putnam Mathematical Competition and thus represents university-level proof tasks well beyond routine textbook exercises. Among many reasons, those failures can be partly traced to the fact that theorem proving offers an \emph{exceptionally sparse reward landscape}: an LLM receives little or no feedback until a full proof (or occasional subgoal) is completed. Most action sequences yield zero signals, making it difficult to learn valuable heuristics or to guide exploration effectively. Overcoming this sparsity while preserving formal soundness remains a central challenge, underscoring the need for strategies that can combine structured search and principled reward shaping.

\begin{wrapfigure}{r}{0.55\textwidth}
  \begin{center}
    \includegraphics[trim={1em 1em 1em 1em}, clip, width=0.55\textwidth]{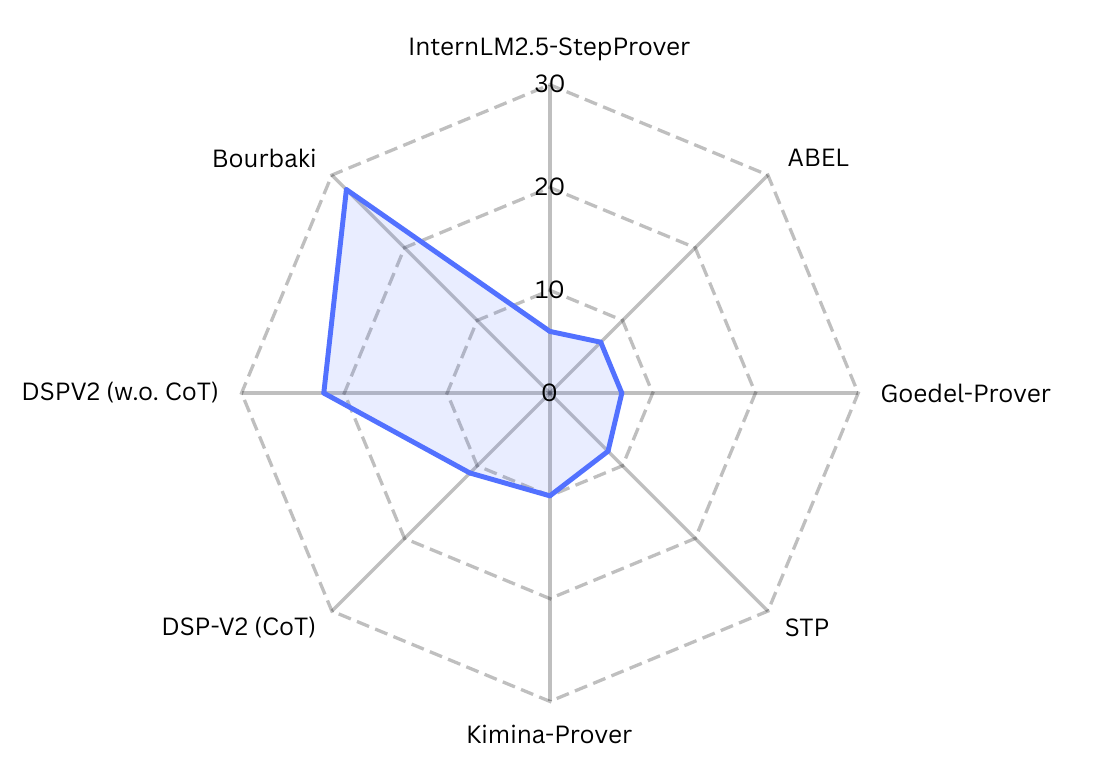}
  \end{center}
  \caption{Comparison of Bourbaki with the strongest existing 7B models on the \texttt{PutnamBench} benchmark, demonstrating a new state-of-the-art at this scale. DSP-V2 (w/o CoT) and DSP-V2 (CoT) refer to the DeepSeek-Prover-v2 7B model evaluated without and with chain-of-thought prompting, respectively.}
\end{wrapfigure} 

One promising direction is to take inspiration from how human mathematicians approach complex proofs. Rather than searching blindly for a complete solution, they typically decompose a problem into a sequence of smaller, intermediate goals that make the final theorem more approachable. This structuring not only organises their reasoning but also provides a kind of internal feedback: proving a useful intermediate step offers a clear signal of progress, even if the overall proof is still incomplete. Mimicking this process in LLM-based proving systems could help address the sparse-reward problem by introducing a denser, more meaningful structure during proof search.

\paragraph{Contributions:} It is natural to formalise this setting using goal-conditioned reinforcement learning (GCRL), where the agent is rewarded for reaching specific subgoals en route to a final objective. However, standard GCRL assumes that the set of goals is given in advance. In contrast, our setting requires the agent to generate its subgoals during the proof process. To capture this, our first contribution is the introduction of a new variant of Markov decision processes (MDPs), which we call self-generated goal-conditioned MDPs or sG-MDPs for short. sG-MDPs allow agents to generate their subgoals dynamically, conditioned on the history of interactions (e.g., previously proven subgoals or applied Lean tactics), making them well-suited for theorem-proving settings where each new discovery or partial result can naturally shift the focus toward a new set of relevant subgoals. In our case, these subgoals are proposed by an LLM and correspond to valid Lean statements that must themselves be formally proven by generating relevant Lean4 tactics. With this formulation in place, we apply search strategies to navigate the space of possible proof trajectories. In particular, we can leverage Monte Carlo Tree Search (MCTS) to explore and expand subgoal candidates, but the framework is general and training with reinforcement learning solvers is also possible, as we detail in Section \ref{Sec:Exp}. 

We call our algorithm Bourbaki, inspired by the pseudonymous group of French mathematicians known for their rigorous and systematic approach to mathematics. Bourbaki is designed to support ensembling multiple LLMs, allowing it to benefit from diverse model behaviours and strengths. In our initial implementation, we ensemble DeepSeek-Prover-v2–7B and Kimina-7B, which we denote as Bourbaki (7B), but the framework is modular, and other models can be easily integrated. We evaluate Bourbaki (7B) on the challenging \texttt{PutnamBench} benchmark. Bourbaki (7B) can solve 26 problems, marking a substantial improvement over the previous 7B state-of-the-art model, Kimina-7B, which solved only 10 problems as reported on the official leaderboard \footnote{https://trishullab.github.io/PutnamBench/leaderboard.html}. Our model also outperforms other strong 7B-scale baselines, including InternML-7B, DeepSeek-Prover-v2 with and without chain-of-thought reasoning, demonstrating the effectiveness of subgoal-guided search and self-generated goal conditioning in tackling competition-level mathematical reasoning.

\section{Related Work}
Automated theorem proving in formal languages (e.g. Lean, Isabelle) is a grand challenge in AI, demanding complex reasoning and extensive domain knowledge. 
RL has emerged as a key paradigm to train LLMs for theorem proving, often through expert iteration (alternating proof generation and learning) or self-play techniques. InternLM2.5-StepProver \citep{wu2024internlm2} is an RL-based prover that scales up expert iteration on Lean using a learned critic model guiding the search: at each iteration, the critic selects relatively easier remaining problems for the prover to attack, which leads to discovering deeper proofs more efficiently.  This targeted search mitigates the high failure rate typical in brute-force exploration. 

Where InternLM2.5 tackled theorem proving with vast compute and data, ABEL \citep{gloeckle2024abel} explores the opposite extreme: \textit{how far can we get with minimal training data?}. ABEL introduces a scalable and computationally efficient online RL framework that serves as a strong baseline with only a few hundred training examples. The approach combines online RL with Lean’s proof environment without relying on large offline proof corpora. Despite the tiny dataset, ABEL’s carefully designed training (inspired by hyper-tree proof search) achieved results comparable to state-of-the-art provers that use orders of magnitude more data. ABEL demonstrates that with an optimised RL pipeline (including techniques like MCTS and careful state representation), even a sparsely trained prover can reach near-SOTA performance.

A different paradigm is explored by STP (Self-play Theorem Prover) proposed by \citep{dong2025stp}. STP addresses the data scarcity problem by generating new training data on the fly through self-play between two roles: a conjecturer and a prover. In STP, the conjecturer LLM proposes novel conjectures (new theorem statements) that are \textit{just beyond the prover’s current ability}, while the prover LLM tries to prove both the original training theorems and these new conjectures using expert iteration. The conjecturer is trained iteratively on those conjectures that the prover barely managed to prove, thus curriculum learning increasingly challenging tasks over time. This loop enables indefinite self-improvement without needing external data, as the system manufactures its own harder problems in a controlled way.

While the above works focus on improving RL given limited data, Goedel-Prover \citep{lin2025goedel} takes a complementary approach: \textit{expand the training data itself via automated formalization and iterative proving} The authors identify the lack of formal training data as a key bottleneck and tackle it head-on by building two large datasets from scratch. First, they use LLMs to perform auto-formalization: converting natural language math problems (from the Numina dataset) into equivalent formal statements in Lean. Next, they employ a sequence of provers (trained iteratively) to accumulate proofs for as many of these statements as possible. Each prover in the sequence learns from the proofs found by previous ones and then attempts new problems, an iterative bootstrapping reminiscent of expert iteration on a grand scale. Through this process they construct Goedel-PSet-v1-solved, a dataset containing formal proofs for over 0.8 M statements. Using this corpus of data, the authors apply RL fine-tuning, pushing miniF2F performance above 60 \% (Pass@32). This offers a different yet complementary path to advance formal reasoning, highlighting that more data can be as powerful as more sophisticated algorithms.

Kimina-Prover \citep{wang2025kimina} introduces a large-scale RL-trained theorem prover based on a 72B-parameter foundation model. The authors employ a “reasoning-driven” RL training paradigm to specialise it for formal proof tasks. A key idea is to instil a \textit{structured formal reasoning pattern} into the model. During training, Kimina-Prover learns to emulate human problem-solving strategies by iteratively generating and refining proof steps, rather than relying on blind search or single-step guesses. This structured pattern (called a formal reasoning pattern in the paper) guides the model to construct proofs step-by-step in a logical manner, which improves reliability and sample efficiency.  The result of this large-scale RL pipeline is a new state-of-the-art on miniF2F: Kimina-Prover achieves 80.7\% success (Pass@8192) on the miniF2F benchmark. This work underscores the value of combining very large LLMs with RL tailored to structured reasoning: by leveraging scale, pattern-based training, and ample compute, Kimina-Prover sets a new benchmark for what open models can achieve in theorem proving.

DeepSeek-Prover \citep{ren2025deepseek} explores integrating informal reasoning into formal proof training via RL. DeepSeek-Prover-V2 is built as an open-source Lean 4 prover that introduces a sub-goal decomposition technique to bootstrap its training. in a cold-start phase, DeepSeek-V3 is prompted to break down complex problems into a sequence of simpler sub-goals, and attempt to solve those sub-goals. The successful sub-goal proofs, along with DeepSeek-V3’s step-by-step reasoning traces, are then combined into a chain-of-thought style dataset for training DeepSeek-Prover-V2 model.
This provides an initial training signal that mixes informal reasoning insights with formal proof steps, jump-starting the RL training for DeepSeek-Prover-V2. Using this hybrid initialisation, the team then performs reinforcement learning (and further fine-tuning) to train a unified model capable of formal proofs. The final model successfully solves 49 out of 658 problems from PutnamBench, setting a new state-of-the-art among language models with over 100 B parameters.

Beyond purely RL-based methods, recent progress in automated theorem proving has increasingly emphasised the use of Monte Carlo Tree Search (MCTS) in combination with language models to more effectively navigate complex proof spaces. \cite{AlphaProof,xin2024deepseekproverv15harnessingproofassistant}.  \citep{han2022proofartifactcotrainingtheorem} introduces the Proof Artefact Co-Training (PACT) framework for the Lean theorem prover, leveraging transformer-based models trained jointly on tactic sequences and proof terms. Although PACT used a uniform goal-expansion strategy in contrast to MCTS, it established a foundation for data-efficient tactic prediction and highlighted the limitations of breadth-first search-type strategies in high-dimensional proof environments.

The work in \citep{lample2022hypertreeproofsearchneural} explicitly adapts AlphaZero-style MCTS to theorem proving in Lean formal language. Recognising that Lean tactics often generate multiple sub-goals, HTPS extended traditional MCTS to operate on hypergraphs rather than simple trees and achieved a significant performance boost on the \text{miniF2F} benchmark. Extending this progress, \citep{wang-etal-2023-dt} introduces the Dynamic-Tree Solver (DT-Solver),   addressing the inefficiencies of fixed-resource allocation in tree search. By estimating the difficulty of proof states through a learned value function, DT-Solver dynamically distributes the search effort, allocating more simulations to uncertain or complex nodes. This adaptive strategy improves performance and underscores the importance of exploration-aware MCTS methods in formal reasoning.

Leveraging these advances, \citep{aniva2025pantograph} developed Pantograph,  a Lean 4 proof interaction interface designed to build more versatile, automated theorem provers. Pantograph provides APIs to execute tactics, track dependencies, and manage goal states, while resolving Lean-specific challenges such as metavariable coupling. By decoupling sub-goal execution and enabling state backtracking, Pantograph facilitates the integration of MCTS and other search strategies in Lean without compromising logical soundness.

\section{Self-generated Goal-conditioned Markov Decision Process}\label{sec:formulation}
We start by introducing the well-known goal-conditioned Markov Decision Process \citep{pmlr-v37-schaul15,liu2022goalconditionedreinforcementlearningproblems} as a tuple $\left( \mathcal{S}, \mathcal{A}^-, T, R, \mathcal{G} \right)$ where $\mathcal{S}$ is a set of states, $\mathcal{A}^-$ is a set of actions, $\mathcal{G}$ is a set of goals, $T : \mathcal{S} \times \mathcal{A}^- \times \mathcal{G} \to \mathcal{S}$ is a transition function and $R : \mathcal{S} \times \mathcal{A}^- \times \mathcal{G} \to \mathbb{R}$ is a reward function. 
It is usually used to find a policy that would be good for a set of different goals. In other words, in goal-conditioned Markov decision processes, we wish to find a policy $\pi^{\star}$ that maximises: 
\begin{equation*}
    \pi^{\star} \in \arg\max_{\pi} \mathbb{E}_{\textbf{s}_0, g \sim \mathcal{P}_{\mathcal{G}},{a}_t,{s}_{t+1} } \left[\sum_{t} \gamma^{t} {R}(s_t, a_t, g)\right],
\end{equation*}
where $\gamma \in (0,1]$ is a discount factor, $\textbf{s}_0$ the initial state, ${s}_{t+1} \sim \mathcal{T}(\cdot|,s_t, a_t)$ the successor state, and $\mathcal{P}_{\mathcal{G}}$ the distribution over possible goals. Moreover, the action $a_t$ is now sampled from a goal-conditioned policy such that $a_t \sim \pi(\cdot|s_t, g)$.

While goal-conditioned MDPs are well-studied in the literature, they typically assume a predefined set of goals or a fixed distribution over them. In our setting, however, such assumptions are too restrictive—when attempting to prove a new theorem, we cannot assume prior knowledge of which subgoals will be useful. Instead, we require the agent to dynamically propose its own sub-goals during the proving process, based on the evolving proof state. 

\paragraph{Self-Generated Goal-Conditioned MDPs:} We formulate this problem using a new paradigm: self-generated goal-conditioned Markov Decision Process (sG-MDP). Here, we augment goal-conditioned MDP with an extra tuple $\left(\mathcal{A} = \mathcal{A}^+ \cup \mathcal{A}^-, T_\mathcal{G}, \text{IsPrimitiveAction}, \text{IsGoal}, \text{ToGoal}, \text{Solves} \right)$ where $\mathcal{A}^+$ are additional action that can create subgoals, the transition function $T_\mathcal{G} : \mathcal{S} \times \mathcal{A} \times \Delta(\mathcal{G}) \to \mathcal{S} \times \Delta(\mathcal{G})$ also acts on goals, with $\Delta(\mathcal{G})$ denoting the set of all possible ordered sets of goals, and where the agent can create a new sub-goal at any point in the trajectory with $\text{IsPrimitiveAction} : \mathcal{S} \times \mathcal{A} \to \{\text{True}, \text{False}\}$ checking if an action is a valid primitive that acts on the state, $\text{IsGoal} : \mathcal{A} \to \{\text{True}, \text{False}\}$ checking if an action is a valid new sub-goal, $\text{ToGoal} :  \mathcal{A} \to \mathcal{G}$ converting valid goals from the action space to the goal space and $\text{Solves} : \mathcal{S} \times \mathcal{A} \times \mathcal{G} \to \{\text{True}, \text{False}\}$ checking if a primitive action can solve a goal in a given state.
We assume that when IsGoal is True, IsPrimitiveAction would be False.
The transition function has the following form $T_\mathcal{G}(s_t, a_t, g_t)$:
\begin{align*}
    s_{t+1}, g_{t+1} \sim \begin{cases}
        s_t, g_t & \text{if } \neg \text{ IsPrimitiveAction}(s_t, a_t) \land \neg \text{ IsGoal}(a_t) \\
        s_t, g_t \cup \text{ToGoal}(a_t) & \text{if } \text{ IsGoal}(a_t) \\
        T(s_t, a_t, \bar g_t), \text{RemoveLast}(g_t) & \text{if } \text{ IsPrimitiveAction}(a_t) \land \text{ Solves}(s_t, a_t, \bar g_t) \\
        T(s_t, a_t, \bar g_t), g_t & \text{if } \text{ IsPrimitiveAction}(a_t) \land \neg \text{ Solves}(s_t, a_t, \bar g_t) \\
    \end{cases}
\end{align*}
where $\bar g_t$ denotes the last element in the ordered set of goals $g_t$ and RemoveLast returns the same stack without its last element.
The policy $\pi(a_t  | s_t, \bar g_t)$ is always defined over the last goal of the stack.
This new formulation allows us to define a smoother reward function depending on Solves$(s_t, a_t, \bar g_t)$.

We aim at solving the following optimisation problem:
\begin{align}
    \underset{\pi}{\arg\, \max\,} \mathbb{E}_{s_0, g_0 \sim D, a_t \sim \pi(\cdot| s_t, \bar g_t), s_{t+1}, g_{t+1} \sim T_\mathcal{G}(s_t, a_t, g_t)}\Big[\sum_t^\infty \gamma^t (R(s_t, \bar g_0) + \lambda R(s_t, \bar g_t))\Big]
\end{align}
where $\lambda$ controls the trade-off between solving the initial goal and solving intermediate sub-goals.

\subsection{Tokenised Proof Search}

We now present how proof search in Lean can be an instantiation of sG-MDPs.
We introduce an additional vocabulary set $\mathcal{V}$ to describe the space of all possible tokens that are used to represent a proof.
This vocabulary set is the usual token space of a given LLM.
In this context, the state space $\mathcal{S}$ consists of the proof states, where each state $s_t$ is defined as a sequence of tokens from $\mathcal{V}$ that make up the proof.
The action space $\mathcal{A}^-$ is the set of all possible combinations of all tokens in the vocabulary.
The set of goals $\mathcal{G}$ represent all the possible conjectures or theorems that can be defined in Lean.
The initial goal $\bar g_0$ is given by the current theorem we aim to prove.
The transition function $T$ concatenates the tokens composing the action into the new state.
The reward function is defined as 1 if the goal is solved and 0 otherwise.

To make it an sG-MDP, we also introduce the needed extra tuple.
Some Lean tactics can introduce explicit intermediate statements or lemmas, which we refer to as \emph{conjectures}. These conjectures -- often constructed using \texttt{have} statements or auxiliary definitions -- can be verified independently in Lean and are not required to complete the original goal.
Those are used to construct subgoals in the sG-MDP.
Because the initial action space is already expressive enough, we have $\mathcal{A} = \mathcal{A}^- \cup \mathcal{A}^+ = \mathcal{A}^-$ which is still the set of all possible combinations of all tokens in the vocabulary.
The function IsPrimitiveAction checks if the tokens composing the action are a valid Lean tactic in the current state of the proof.
The function IsGoal checks if the tokens composing the action are a valid Lean \emph{conjecture}.
We want to highlight that although only a limited number of possible actions or goals are valid in a state, we do not assume to know those beforehand.
The function ToGoal is an identity function over the set of all possible combinations of all tokens.
The function Solved checks when a goal is solved. Note that it is already defined within the reward function of our MDP.

Compared to the traditional tree search method based on LeanDojo \citep{yang2023leandojo}, which does not allow for creating subgoals, we rely on PyPantograph \citep{aniva2025pantograph} to compute self-goal conditioned MDP functions at any stage of the proof.
By formulating Lean proof search as an sG-MDP, we can leverage the framework to guide the search for proofs, generate subgoals, and provide a denser reward structure that encourages the discovery of verifiable conjectures.

\subsection{Self-Generated Goal-Conditioned MCTS}\label{sec:mcts}

To solve the previously described sG-MDP, we choose to rely on Monte-Carlo Tree Search (MCTS).
Unlike prior work, which relies solely on terminal success as the reward for MCTS, we augment the reward by measuring how many locally introduced conjectures can be independently verified, without any additional pretrained critic. 

The tree search is guided by the policy \(\pi(s_t, \bar g_t)\) that samples tactics and subgoals, and a value function that is computed entirely through verified outcomes -- without relying on any learned critic. %
For each node in the tree representing a different \((s_t, g_t)\), we maintain a visit count \(N(s_t, g_t)\) and an accumulated value \(W(s_t, g_t)\), used to compute an empirical mean \(\bar{V}(s_t, g_t) = W(s_t, \bar g_t)/N(s_t, g_t)\).

\paragraph{Selection:} The search begins with a tree having a single initial root \((s_0, g_0)\). 
The selection step involves choosing the next node to expand based on the Upper Confidence Bound Applied to Trees (UCB) algorithm, which balances exploration and exploitation. We select the node with the highest UCB score, given by:
\[ \bar{V}(s_t, g_t) + C \cdot \sqrt{\frac{\log N(s_t, g_t)}{N(s_t, g_t)}} \]
where \(C\) is a hyperparameter controlling the trade-off between exploration and exploitation.

\paragraph{Expansion:} Once a node is selected, we query the policy model \(\pi(s_t, \bar g_t)\) to suggest tactic candidates with a maximum of N trials.
As soon as a tactic is validated by PyPantograph, we add the new node in the tree and perform the estimation step on it.
If no valid tactic is found within the N trials, we prevent this node from being selected again and return to the selection phase.
We also exclude the node having already $N$ children from the selection process.

\paragraph{Estimation:}
The initial value of a new node is set to $R(s_t, \bar g_t)$. 
We propose multiple ways to design the reward function in the experiment part.
It can promote the number of solved conjectures, the depth of the proof, etc.

\paragraph{Back-propagation:} 
The backup step involves updating the visit counts and accumulated values of all nodes in the tree, from the expanded node to the root. We use the following update rules:
\[ N(s_t, \bar g_t) \leftarrow N(s_t, \bar g_t) + 1 \]
\[ W(s_t, \bar g_t) \leftarrow W(s_t, \bar g_t) + R(s_t, \bar g_t) \]
where \(R(s_t, \bar g_t)\) is the reward obtained by evaluating the state \((s_t, \bar g_t)\).

The search process continues until a termination condition is met, such as a maximum number of iterations $K$ or a satisfactory proof being found. The final proof is then extracted directly from the tokens in $s_t$.

\section{Results} \label{Sec:Exp}
\paragraph{Experimental Setup:}  
We evaluate our method on the challenging \texttt{PutnamBench} dataset\footnote{The dataset includes 658 Putnam Competition problems in Lean 4. The \texttt{PutnamBench} benchmark is continuously updated. Some earlier results reported in prior work were based on the 644-problem version} by measuring the number of correctly completed proofs. All experiments are conducted in the Lean 4 environment and Pantograph for tactic validation and goal verification. We ensemble DeepSeek-Prover-v2–7B~\cite{ren2025deepseek} and Kimina-7B~\cite{wang2025kimina} as the base policy model in our main results, and use vLLM for efficient batch generation on the base models. A proof is considered to be solved successfully if it completes the goal and type-checks in the Lean REPL. We report pass@$k$ under matched sample budgets to ensure a fair comparison with current state-of-the-art 7B and 8B models, including both tree search methods~\citep{wu2024internlm2, gloeckle2024abel} and whole proof generation method~\citep{lin2025goedel, dong2025stp, wang2025kimina, ren2025deepseek}. For each run, we allow up to $N=10$ tactic candidates per node, with a maximum number of iterations of $K=512$.

\paragraph{Implementation Details:}
We implement the goal-conditioned framework using \texttt{Pantograph v0.3.2}, with \texttt{Lean v4.20.1} and \texttt{mathlib v4.20.1} serving as backends for tactic validation and for verifying solved goals and sub-goals in given proof states at each step. In practice, the sG-MDP framework can be naturally addressed by any tree search or MCTS variant. In our implementation, we estimate the value of each proof state using a combination of depth-based metrics and the number of solved conjectures, which provides denser feedback signals during the search before completing the entire proof. Other value-estimation methods can be seamlessly integrated into our formulation, for instance, prior work employs \texttt{Leandojo} \citep{yang2023leandojo} or pretrained critics \citep{wu2024internlm2} to estimate state values and guide the search process. Although pretrained critics and PRMs are widely used in contemporary tree-search methods, they remain challenging to train effectively. We therefore continue to explore alternative value-function formulations that leverage Lean-verifier feedback to enhance reasoning capabilities.

We also observed that some policy models generate ``heuristic'' tactics -- such as \texttt{apply?} from \texttt{mathlib4} -- which return a set of candidate tactics rather than a single fixed one. Therefore, these tactics produce an information set rather than a deterministic proof state. In our implementation, we handle such ``heuristic'' tactics by using the plain base model to complete the remainder of the proof. While this approach introduces diversity in possible solutions, we are exploring prioritising static proof or validating individual tactics during tree search, thereby improving the soundness and reliability of the resulting proofs.

\begin{table}[ht]
\centering
\caption{Comparison with state-of-the-art models on the \texttt{Putnambench} dataset. The tags CoT and non-CoT refer to two generation modes of a unified model, each guided by a different prompt.}
\label{tab:putnam-comparison}
\resizebox{0.95\textwidth}{!}{%
\begin{tabular}{@{}llccc@{}}
\toprule
\textbf{Method} & \textbf{Model size} & \textbf{Sample budget} & \textbf{\# Solved theorems}\\
\midrule
\multicolumn{4}{@{}l}{\textit{Tree Search Methods}} \\
\midrule
InternLM2.5-StepProver~\citep{wu2024internlm2} & 7B & $256 \times 32 \times 600$ & 6/644 \\
ABEL~\citep{gloeckle2024abel} & 8B & 596 & 7/644 \\
\midrule
\multicolumn{4}{@{}l}{\textit{Whole-proof Generation Methods}} \\
\midrule
Goedel-Prover-SFT~\citep{lin2025goedel} & 7B & 512 & 7/644\\
STP~\citep{dong2025stp} & 7B & 64 & 7/644 \\
& & 3200 & 8/644\\
Kimina-Prover-Preview-Distill~\citep{wang2025kimina} & 7B & 192 & 10/644 \\
DeepSeek-Prover-V2 (non-CoT)~\citep{ren2025deepseek} & 7B & 128 & 15/658 \\
 &  & 1024 & 23/658 \\
DeepSeek-Prover-V2 (CoT)~\citep{ren2025deepseek}& 7B & 128 & 10/658 \\
 &  & 1024 & 11/658\\
 \midrule
\textbf{Bourbaki (7B)} & 7B & 512 & \textbf{26/658}\\
\bottomrule
\end{tabular}
}
\end{table}

\paragraph{Main Results:}  
We present our results on the \texttt{PutnamBench} dataset, comparing our approach against recent 7B and 8B provers across both tree search and whole-proof generation methods, as shown in Table~\ref{tab:putnam-comparison}. Bourbaki establishes a new state-of-the-art for 7B models on \texttt{PutnamBench}, solving 26/658 theorems at pass@512 -- outperforming the previous best of 23/658 solved theorems at pass@1024 by DeepSeek-Prover-V2-7B and the previous 7B state-of-the-art model on the \texttt{Putnambench} leaderboard Kimina-7B with 10/644 solved theorems. Our method completes more proofs with fewer samples, and notably, it discovers some new theorems at pass@512 that are not found at pass@1024 using the base models -- demonstrating the strength of our goal-conditioned formulation in enhancing reasoning capabilities.

To further demonstrate our sample efficiency, we apply the sG-MDP framework on top of the STP prover at pass@64 and pass@128. By incorporating our conjecture-aware actions, our method generates one additional correct proof in both settings, and notably discovers a proof at pass@128 that only appears at pass@3200 in the base STP model. This improvement highlights the advantages of structured exploration and intermediate conjecture signals in maximising the value of each model query. As shown in Table~\ref{tab:mcts-efficiency}, this type of improvement generalises across both base models: we compare performance with and without Bourbaki using STP at pass@64 and pass@128, and DeepSeek-Prover-V2 (non-CoT) at pass@128. Our method consistently improves proof success rates while maintaining the same sample budget.

Beyond improvements in the number of solved proofs, our method enhances proof diversity in the solved theorems. For each solved theorem, it generates more diverse tactics and produces more varied proofs compared to the base model. For example, at pass@128 on \texttt{putnam\_1990\_a1}, DeepSeek-V2-7B generates 1 correct proof, while Bourbaki produces 4 correct proofs; on \texttt{putnam\_2001\_a1}, DeepSeek-V2-7B generates 36 correct proofs, whereas Bourbaki produces 105 at pass@128.

These results demonstrate the effectiveness of our sG-MDP framework in guiding proof search through dynamically generated subgoals. By combining structured exploration with intermediate conjectures, our method helps the prover focus on more promising proof paths under limited sample budgets. This leads not only to solving more problems, but also to generating a greater variety of correct proofs in solved theorems. The consistent improvements across different base models show that self-generated goal conditioning provides a generalizable mechanism for enhancing deductive reasoning in formal mathematical problem-solving.
\begin{table}[ht]
\centering
\caption{Sample efficiency demonstration of Bourbaki under small sample budgets using two base models, STP and DeepSeek-v2-7B}
\label{tab:mcts-efficiency}
\begin{tabular}{@{}lcc@{}}
\toprule
\textbf{Method} & \textbf{Sample Budget} & \textbf{\# Solved theorems} \\
\midrule
STP & 64 & 6/644 \\
Ours (Bourbaki on STP) & 64 & 7/644 \\
\midrule
STP & 128 & 7/644 \\
Ours (Bourbaki on STP) & 128 & 8/644 \\
\midrule
DeepSeek-Prover-v2 (non-CoT) & 128 & 15/658 \\
Ours (Bourbaki on Deepseek-Prover-v2) & 128 & \textbf{23/658} \\
\bottomrule
\end{tabular}
\end{table}

\section{Conclusion}
This paper introduces a novel framework for automated theorem proving, leveraging self-generated goal-conditioned Markov Decision Processes (sG-MDPs) to tackle the challenges of sparse rewards and long proof horizons. By dynamically generating subgoals and incorporating intermediate conjectures, the proposed approach enhances the efficiency and effectiveness of proof search. The Bourbaki system, which ensembles multiple large language models for subgoal generation and tactic synthesis, achieves state-of-the-art results on the PutnamBench benchmark with 7B models, solving 26 problems and outperforming other strong 7B-scale baselines. The results demonstrate the potential of sG-MDPs in guiding proof search and improving deductive reasoning in formal mathematical problem-solving, paving the way for further research and development in this area.

\bibliography{iclr2025_conference}
\bibliographystyle{iclr2025_conference}

\end{document}